\documentclass{article}
\usepackage{stywhispers,amsmath,epsfig}
\usepackage{booktabs}
\usepackage{tabularx}
\usepackage{wrapfig}
\usepackage{graphicx}
\usepackage{subfig}
\usepackage{booktabs}
\usepackage{afterpage}
\usepackage{rotating}
\usepackage[]{siunitx}
\usepackage{url}

\title{Spectral Angle Based Unary Energy Functions for Spatial-Spectral Hyperspectral Classification using Markov Random Fields}
\name{Utsav B. Gewali~\textsuperscript{1} and Sildomar T. Monteiro~\textsuperscript{1,2}}
\address{Chester F. Carlson Center for Imaging Science\textsuperscript{1}, Dept. of Electrical Engineering\textsuperscript{2}\\
		 Rochester Institute of Technology, Rochester, NY\\
		 ubg9540@rit.edu}
\begin{document}
%
\maketitle
\begin{abstract}
In this paper, we propose and compare two spectral angle based approaches for spatial-spectral classification. Our methods use the spectral angle to generate unary energies in a grid-structured Markov random field defined over the pixel labels of a hyperspectral image. The first approach is to use the exponential spectral angle mapper (ESAM) kernel/covariance function, a spectral angle based function, with the support vector machine and the Gaussian process classifier. The second approach is to directly use the minimum spectral angle between the test pixel and the training pixels as the unary energy. We compare the proposed methods with the state-of-the-art Markov random field methods that use support vector machines and Gaussian processes with squared exponential kernel/covariance function. In our experiments with two datasets, it is seen that using minimum spectral angle as unary energy produces better or comparable results to the existing methods at a smaller running time. 
\end{abstract}
\section{Introduction}
\label{sec:intro}
Hyperspectral classification is the process of identifying the material present under each pixel in a hyperspectral image. This is possible as the fraction of incident light reflected by a material at different wavelengths (the spectrum), captured at each pixel of a hyperspectral image, is dependent on the chemical structure of the material. Statistical methods have been successful in predicting the material class from the spectrum~\cite{lu2007}. Traditionally, pixel-wise classifiers were trained to predict the material under a pixel using only the spectrum captured at that pixel. However, since, the materials in a scene are typically distributed in homogeneous regions and the presence of one material can influence the likelihood of another material being present in its vicinity, it has been seen that the classification performance can be significantly  improved by utilizing the spatial information along with the spectral information~\cite{fauvel2013}. 

There have been basically two approaches to build spatial-spectral hyperspectral classifiers. One is to use spatial-spectral features~\cite{benediktsson2005,chen2015}, and the other is to use Markov random fields~\cite{tarabalka2010,liao2015}. In this paper, we explore the use of Markov random field for spatial-spectral classification. Currently, the common classifiers used with Markov random fields are logistic regression~\cite{li2012}, probabilistic support vector machines~\cite{tarabalka2010} and Gaussian processes~\cite{liao2015}. In this paper, we experiment using the exponential spectral angle mapper kernel/covariance function with the support vector machine and the Gaussian process in these methods, and also experiment with combining the the spectral angle mapper, possibly the simplest pixel-wise classifier, with the Markov random field.

\section{Background}
\label{sec:bg}

\subsection{Markov Random Fields}
Markov random fields (MRFs) can be used to exploit the strong dependencies between the neighboring pixels in a hyperspectral image to improve the classification performance. MRFs define a joint probability distribution over all the pixel labels in an image as
\begin{equation}
p(\mathbf{y}) = \frac{1}{Z} \exp \left( -E\left(\mathbf{y}\right) \right),  
\end{equation}
where $\mathbf{y} = \left[y_1,...,y_N\right]^T$ is a vector containing all the $N$ pixel labels in an image, $E\left(\mathbf{y}\right)$ is the total energy of the pixel labels and $Z$ is a normalization constant, $Z = \sum_{\mathbf{y}} \exp \left( -E\left(\mathbf{y}\right) \right)$. The inference about the pixel labels, $\mathbf{y}$, is performed by maximum likelihood estimation, which is equivalent to minimizing the total energy, $E\left(\mathbf{y}\right)$. The energy minimization can be performed by methods like GraphCuts~\cite{boykov2001}. The total energy of the grid-structured Markov random field used for image classification consists of two parts as
\begin{equation}
E\left(\mathbf{y}\right) = \sum_{i \in V} E_i\left(y_i\right) + \sum_{(i,j) \in D} E_{ij}\left(y_i,y_j\right),
\end{equation}
where $E_i\left(y_i\right)$ is the unary energy of the i\textsuperscript{th} pixel with label $y_i$ and $E_{ij}\left(y_i,y_j\right)$ is the pairwise energy between the two neighboring i\textsuperscript{th} and the j\textsuperscript{th} pixels having labels $y_i$ and $y_j$ respectively. V is the set of all the pixels and D is the set of all the edges between 4-neighboring pixels in the image. The unary energy incorporates the spectral information, while the pairwise energy incorporates the spatial information. The unary energy at a pixel i when $y_i=c$ can be defined to be the negative logarithm of the probability that the pixel belongs to the class c, $E_i\left(y_i=c\right) = - \ln \left( P\left(  y_i=c \mid \mathbf{x}_i \right) \right)$. The MRFs with the logistic regression, the support vector machines and the Gaussian process use this energy function in our experiments.  We introduce the unary energy function used with the spectral angle mapper in Section \ref{sec:sam_mrf}. The Potts model was utilized as the pairwise energy function in this paper. It is defined as 
\begin{equation}
E_{ij}\left(y_i,y_j\right) = 
\begin{cases}
    0,& \text{if } y_i = y_j\\
    \beta,              & \text{otherwise},
\end{cases}
\end{equation}
where $E_{ij}\left(y_i,y_j\right)$ is the energy of the edge i-j, when $y_i$ and $y_j$ are the labels of the i\textsuperscript{th} and the j\textsuperscript{th} pixels respectively. $\beta$ is a parameter that represents the cost of the labels $y_i$ and $y_j$  being different, and its value can be learned using cross-validation. 

\subsection{Exponential Spectral Angle Mapper (ESAM) kernel/covariance function}
\label{sec:esam}
The ESAM kernel/covariance function for two inputs $\mathbf{x}_1$ and $\mathbf{x}_2$ is defined as 
\begin{equation}
 k_\text{ESAM}(\mathbf{x}_1,\mathbf{x}_2) = \sigma_0^2  \exp( - \alpha(\mathbf{x}_1,\mathbf{x}_2) / \sigma_1^2 ),
\label{eq:ESAM0}
\end{equation}
where
\begin{equation}
\alpha(\mathbf{x}_1,\mathbf{x}_2) = {\cos}^\text{-1} \left(  \frac{\mathbf{x}_1 \cdot  \mathbf{x}_2 }{\|\mathbf{x}_1\|\,\|\mathbf{x}_2\| } \right),   
\label{eq:ESAM1}
\end{equation}
and, $\sigma_0^2$ and $\sigma_1^2$ are the gain and the scale parameters respectively. $\alpha(.,.) $ is the spectral angle mapper. The parameters are learned from the data while training the models. We introduced this function for biochemical prediction from hyperspectral data with the Gaussian processes in \cite{gewali2016}. A function similar to the ESAM function has been previously used for hyperspectral classification using the support vector machines in \cite{mercier2003}. 

\section{Spectral Angle Mapper-Markov Random Field (SAM-MRF)}
\label{sec:sam_mrf}
The proposed Spectral Angle Mapper-Markov Random Field (SAM-MRF) combines the spectral angle mapper metric and the Markov random field. In this method, the unary potential function at each pixel is defined as the minimum spectral angle between the test pixel and the training spectra belonging to each class. The unary energy at pixel i, when the label $y_i$ is c, is given by 

\begin{equation}
E_i\left(y_i=c\right) = \min_{\mathbf{x}_{tc} \in \substack{\text{ training spectra}\\\text{of class c}}} \alpha\left(\mathbf{x}_i, \mathbf{x}_{tc}\right),
\end{equation}	
where $\mathbf{x}_i$ is the spectrum of pixel i and $\alpha(.,.)$ is the spectral angle mapper from (\ref{eq:ESAM1}). Intuitively, this model introduces a new decision method for determining the class of the pixel from the spectral angle. Unlike the previous methods that only consider the test pixel and make decision by thresholding the spectral angle or choose the class with minimum angle, our approach jointly minimizes the spectral angle and promotes spatial homogeneity across the image. Recently, the study~\cite{tang2015} by Tang et al.\ have combined SAM and MRF using multi-center model and Gaussian normalization, but our method is different in that it directly uses the minimum spectral angle as the unary energy.   

\section{Experiments}
\label{sec:experiments}
We experiment with two publicly available classification datasets: the Indian Pines~\cite{indian_pines2015} and the University of Pavia\footnote{both obtained from \url{http://www.ehu.eus/ccwintco/index.php?title=Hyperspectral_Remote_Sensing_Scenes}}. The Indian Pines dataset  contains a 145 $\times$ 145 hyperspectral image of a 2 $\times$ 2 miles area, covering agricultural land and forest, in Northwest Tippecanoe County, Indiana collected by the Airborne Visible/Infrared Imaging Spectrometer (AVRIS). The pixel diameter is around \SI{4}{\metre} and each pixel contains 220 spectral bands, with wavelengths ranging from \SI{400}{\nano\metre} to \SI{2500}{\nano\metre}. Twenty water absorption bands were removed from the image as pre-processing. In our experiments, only the 14 material classes, each of which were present at 150 or more pixel locations were used. The University of Pavia dataset was collected by Reflective Optics System Imaging Spectrometer (ROSIS) over city of Pavia in northern Italy. It contains 103 bands in visible and near-infrared (\SI{400}{\nano\metre} to \SI{900}{\nano\metre}). The image is 610 $\times$ 340 pixels in size, with each pixel having a diameter of \SI{1.3}{\metre}. There are nine material classes for this image. Full ground truth material cover maps are available for both of the images. Both images are not atmospherically compensated, with the pixels measured in the units of spectral radiance. The spectral radiance in each band of the image were normalized to have a mean of zero and a standard deviation of one.

The pixels in the image were randomly divided into the training set and the testing set. The testing set contained 50 pixels from each class, while the size of the training set was varied from 10 to 70 pixels per class at the increments of 10. 70\% of the training data was used to train the models generating the unary energies, and the remaining 30\% of the training data was used to choose the value of the parameter ($\beta$) in the Potts pairwise energies via cross-validation. The value of $\beta$ was chosen from \{0.01,0.1,1,10,100\} by maximizing the overall accuracy. The unary energies were generated using the logistic regression (LR), the support vector machine (SVM), the Gaussian process (GP) and the spectral angle mapper (SAM). The implementations used are the multivariate logistic regression with L2 regularized weights from LIBLINEAR library~\cite{liblinear}, probabilistic multi-output support vector machine from LIBSVM library~\cite{libSVM}, and the Gaussian process classifiers from the GPML library~\cite{rasmussen2010}. The slack variable and the kernel scale in the SVM was chosen from \{0.001,0.01,0.1,1,10,100,1000\} by training the SVM on 90\% of classifier's training data and validating over the remaining 10\%. The gain of ESAM was set to one while using it with the SVM. GPML library does not contain multi-class classifiers, so binary classifiers were trained in one-vs-one setup and the multi-class probabilities were estimated using the method~\cite{wu2004} by Wu et al. Error function likelihood was used with the GP classifier and the inference was done using Laplace approximation. The hyper-parameters of the covariance function were learned by maximizing the likelihood. The final output labels were produced by Markov random field energy minimization, performed using the graph cut with expansion-move algorithm using the software~\cite{szeliski2008} by Szeliski et al. Overall accuracy over the testing set was used to measure the performance. This procedure was repeated 30 times to produce the mean and the standard deviation of the overall accuracy as the final performance metric.

\begin{table*}[!htb]
\caption{The mean and the standard deviation of overall accuracies as a function of number of training pixels per class for the Indian Pines image. }
\label{table_indian}
\centering

\begin{tabular}{@{}llllllll@{}}
\toprule
 & 10 & 20 & 30 & 40 & 50 & 60 & 70 \\
\midrule
LR & 53.87$\pm$2.3 & 61.62$\pm$2.0 & 64.88$\pm$2.2 & 67.53$\pm$2.5 & 69.28$\pm$2.3 & 70.81$\pm$2.0 & 71.62$\pm$1.8 \\
LR-MRF & \textbf{69.69}$\pm$\textbf{6.2} & \textbf{79.95}$\pm$\textbf{2.9} & 82.53$\pm$2.9 & 83.88$\pm$2.4 & 84.42$\pm$2.6 & 85.10$\pm$2.8 & 85.69$\pm$2.3 \\
SVM-SE & 50.38$\pm$7.3 & 63.04$\pm$5.4 & 69.42$\pm$3.9 & 72.37$\pm$2.7 & 73.41$\pm$4.1 & 76.79$\pm$2.0 & 78.39$\pm$1.7 \\
SVM-SE-MRF & 57.94$\pm$10.9 & 77.19$\pm$6.8 & 82.73$\pm$4.0 & 86.18$\pm$2.4 & 86.61$\pm$4.1 & 89.37$\pm$2.3 & 90.61$\pm$1.9 \\
SVM-ESAM & 47.42$\pm$4.3 & 57.49$\pm$7.4 & 65.64$\pm$4.1 & 68.91$\pm$3.2 & 71.75$\pm$1.9 & 73.71$\pm$2.1 & 74.85$\pm$2.3 \\
SVM-ESAM-MRF & 55.24$\pm$8.2 & 70.55$\pm$10.7 & 81.28$\pm$3.5 & 83.74$\pm$2.6 & 85.57$\pm$2.2 & 87.21$\pm$2.4 & 87.95$\pm$2.3 \\
GP-SE & 51.55$\pm$3.0 & 61.10$\pm$2.2 & 66.79$\pm$2.2 & 70.01$\pm$1.8 & 72.59$\pm$1.9 & 75.60$\pm$2.1 & 77.19$\pm$1.9 \\
GP-SE-MRF & 60.61$\pm$6.8 & 75.89$\pm$4.7 & 81.49$\pm$3.6 & 85.24$\pm$2.2 & 87.31$\pm$2.3 & 88.49$\pm$2.3 & 90.23$\pm$2.2 \\
GP-ESAM & 48.99$\pm$2.7 & 55.96$\pm$2.2 & 60.84$\pm$2.0 & 64.19$\pm$2.6 & 66.43$\pm$2.0 & 69.48$\pm$2.2 & 71.01$\pm$1.8 \\
GP-ESAM-MRF & 59.83$\pm$6.8 & 75.24$\pm$5.6 & 80.53$\pm$2.7 & 82.84$\pm$1.9 & 84.16$\pm$2.0 & 85.99$\pm$2.1 & 86.61$\pm$2.3 \\
SAM & 50.74$\pm$2.2 & 57.46$\pm$2.2 & 60.15$\pm$2.2 & 61.04$\pm$2.1 & 62.97$\pm$2.0 & 63.92$\pm$1.9 & 64.73$\pm$1.8 \\
SAM-MRF & 65.46$\pm$4.7 & 77.97$\pm$3.1 & \textbf{85.22}$\pm$\textbf{3.3} & \textbf{87.02}$\pm$\textbf{2.4} & \textbf{89.28}$\pm$\textbf{2.2} & \textbf{90.88}$\pm$\textbf{1.9} & \textbf{92.00}$\pm$\textbf{1.8} \\
\bottomrule
\end{tabular}
\end{table*}

\begin{table*}[!htb]
\caption{The mean and the standard deviation of overall accuracies as a function of number of training pixels per class for the University of Pavia image.}
\label{table_pavia}
\centering
\begin{tabular}{@{}llllllll@{}}
\toprule
& 10 & 20 & 30 & 40 & 50 & 60 & 70 \\
\midrule
LR & 64.21$\pm$2.8 & 68.24$\pm$1.7 & 70.10$\pm$2.1 & 71.16$\pm$2.3 & 71.72$\pm$2.0 & 72.23$\pm$2.2 & 72.13$\pm$1.6 \\
LR-MRF & 66.01$\pm$2.7 & 71.13$\pm$2.5 & 72.24$\pm$2.8 & 73.41$\pm$2.0 & 74.17$\pm$2.2 & 74.67$\pm$2.2 & 74.53$\pm$2.4 \\
SVM-SE & 69.16$\pm$4.6 & 75.89$\pm$5.9 & 79.16$\pm$5.3 & 83.07$\pm$2.8 & 83.72$\pm$2.4 & 85.85$\pm$2.7 & 86.79$\pm$2.0 \\
SVM-SE-MRF & 68.84$\pm$5.4 & 76.28$\pm$5.5 & \textbf{80.10}$\pm$\textbf{5.8} & \textbf{84.19}$\pm$\textbf{2.9} & \textbf{85.81}$\pm$\textbf{2.3} & \textbf{88.01}$\pm$\textbf{2.3} & \textbf{88.90}$\pm$\textbf{2.1} \\
SVM-ESAM & 66.81$\pm$7.5 & 75.54$\pm$3.6 & 78.01$\pm$4.4 & 79.73$\pm$3.2 & 80.33$\pm$2.9 & 82.47$\pm$2.9 & 83.51$\pm$2.4 \\
SVM-ESAM-MRF & 66.70$\pm$8.0 & 76.19$\pm$3.6 & 79.27$\pm$6.0 & 82.21$\pm$4.1 & 83.73$\pm$3.0 & 85.47$\pm$2.9 & 87.01$\pm$2.5 \\
GP-SE & 73.07$\pm$3.3 & 76.31$\pm$2.2 & 79.18$\pm$2.9 & 81.88$\pm$2.4 & 83.44$\pm$2.2 & 86.03$\pm$1.9 & 87.37$\pm$1.8 \\
GP-SE-MRF & 73.95$\pm$3.6 & 76.91$\pm$2.5 & 79.61$\pm$3.0 & 82.68$\pm$2.5 & 84.30$\pm$2.4 & 87.28$\pm$1.7 & 88.23$\pm$1.8 \\
GP-ESAM & 71.99$\pm$2.3 & 75.79$\pm$2.2 & 77.93$\pm$2.1 & 78.93$\pm$2.1 & 80.14$\pm$1.7 & 81.32$\pm$2.0 & 82.27$\pm$2.1 \\
GP-ESAM-MRF & 72.56$\pm$2.3 & 76.45$\pm$2.5 & 78.81$\pm$2.4 & 79.81$\pm$2.4 & 81.03$\pm$1.9 & 82.50$\pm$2.0 & 83.25$\pm$2.2 \\
SAM & 71.90$\pm$2.7 & 74.60$\pm$2.4 & 76.34$\pm$2.0 & 77.11$\pm$2.2 & 77.47$\pm$2.0 & 78.50$\pm$2.0 & 78.91$\pm$2.1 \\
SAM-MRF & \textbf{74.40}$\pm$\textbf{3.2} & \textbf{77.33}$\pm$\textbf{2.4} & 78.34$\pm$2.5 & 79.80$\pm$2.0 & 79.93$\pm$2.0 & 80.66$\pm$1.2 & 81.53$\pm$2.4 \\
\bottomrule
\end{tabular}
\end{table*}

\begin{figure}
\centering
	 \subfloat[][Ground truth]{\includegraphics{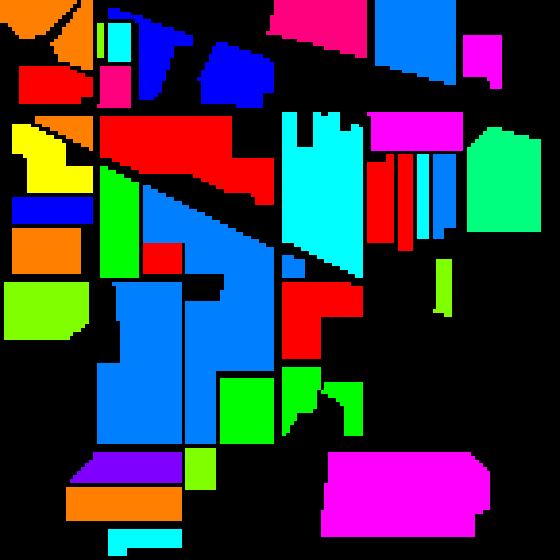}} \\
	 \subfloat[][Training Pixels]{\includegraphics{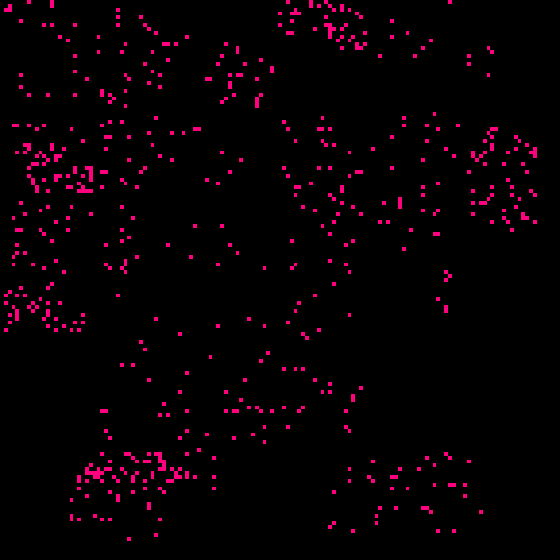}} \\
	 \subfloat[][Material classes]{\includegraphics{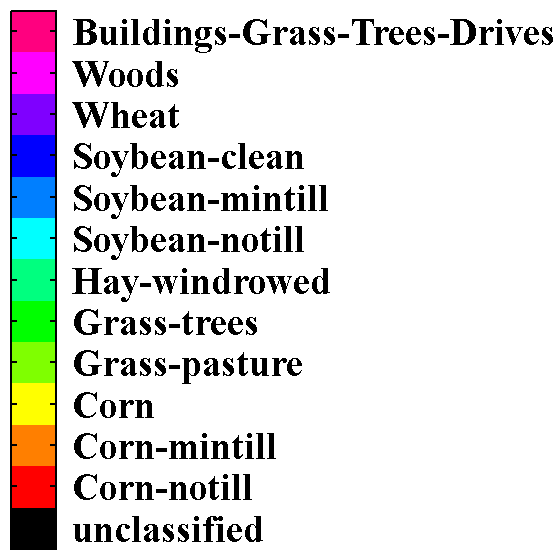}} \\
	 \subfloat[][SAM]{\includegraphics{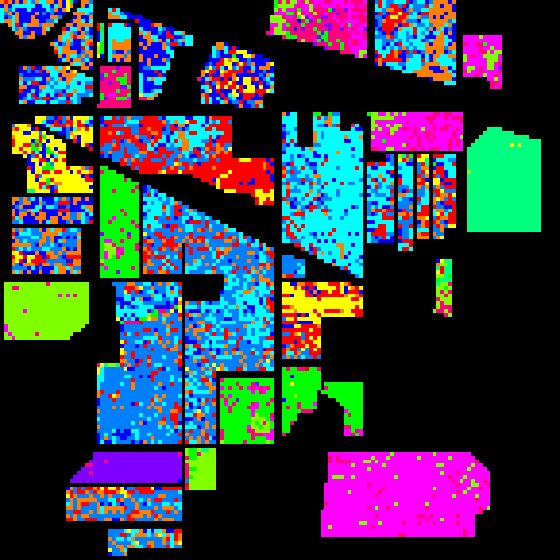}} \\
	 \subfloat[][SAM-MRF]{\includegraphics{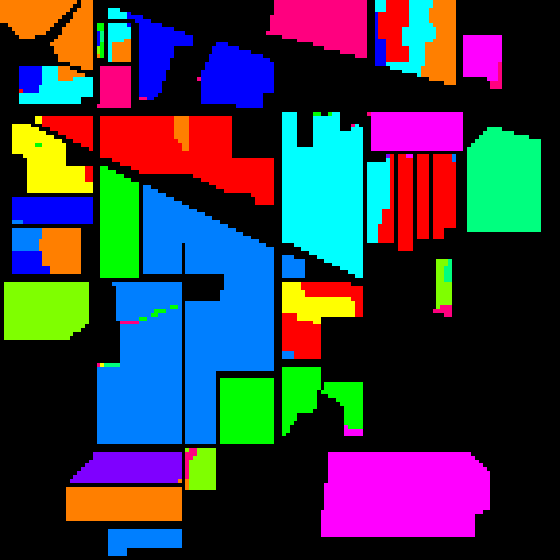}} 
\caption{Classification results on the Indian Pines image.}
\label{fig:indian_pines_images}
\end{figure}
\section{Results}
\label{sec:results}

Tables \ref{table_indian} and \ref{table_pavia} compare the performance of all the methods on the Indian Pines image and the University of Pavia image respectively. The logistic regression, the support vector machine and the Gaussian process and the spectral angle mapper have been denoted as LR, SVM, GP, and SAM respectively. The abbreviation of the name of the kernel/covariance function used with the SVM and the GP has be appended at the end of the methods name. The kernel/covariance function used are the squared exponential function (SE) and the exponential spectral angle mapper (ESAM). Those methods which use Markov random field energy minimization have MRF appended at the end of their name. Figure \ref{fig:indian_pines_images} shows one of the classification maps produced by the proposed, SAM-MRF, when the number of training pixels per class was 50 on the Indian Pines image. When there were 50 samples per class in the training set, the SAM-MRF was the most accurate and took 3.07$\pm$0.2 seconds to compute the classification map. This is much better than 44.8$\pm$1.7 seconds taken by the second most accurate GP-SE-MRF. The second fastest method was LR-MRF, taking 4$\pm$0.3 seconds. 

\section{Discussion}
Compared to the state-of-the-art methods, the SAM-MRF method produced superior accuracies on the Indian Pines image and comparable accuracies on the University of the Pavia image. This could have been due to the two major differences between these datasets. The Indian Pines image contains many large homogeneous areas, and has less distinct material classes, e.g., most classes are the different types of vegetation. Hence, when SAM-MRF is applied to this image, the minimum angle for each class are most likely to be comparable to each other in magnitude as the material classes are less distinct. The Markov random field can then choose an appropriate label from the labels having comparable low spectral angles by considering the neighbors of the pixels which are highly related for this image, rather than just choosing the label having the minimum of the roughly equal spectral angles. The University of Pavia image, on the other hand, has fewer, smaller homogenous areas, and has more distinct classes, such as asphalt, trees, gravel, shadows and paint. Hence, the accuracy after MRF for the University of Pavia image is not improved much and is highly dependent on the pixel-wise classification accuracy, which is poor in case of SAM-MRF  Hence, SAM-MRF performs poorly on this image. This also explain why applying MRF to the University of Pavia image, in general, only increased the accuracy by about 2\% for all the methods. One possible way to improve the classification performance on the University of Pavia dataset could be to use a spatial-spectral features, such as the extended morphological features~\cite{benediktsson2005}, with the proposed methods.    

It was seen that using the ESAM kernel/covariance function did not improve the performance over using the squared exponential kernel/covariance function for both the SVM and the GP, indicating that the spectral angle based functions are not necessarily better for classification when used with these classifiers. The SVM based methods and the GP based methods did show significant difference in performance, however the SVM based ones were significantly faster. In the experiments, even the naive implementation of SAM-MRF was faster than the robust implementations of other methods. This is due to the simplicity of SAM-MRF. SAM-MRF could be made even faster and scalable to very large datasets by using heuristics, e.g., k-d tree nearest neighbor search~\cite{muja2014}, to find the approximate minimum angle.

\label{sec:discussion}

\bibliographystyle{IEEEbib}
\bibliography{whispers2}

\end{document}